\documentclass{vldb}

\usepackage{amsmath}
\usepackage{graphicx}
\usepackage{balance}  
\usepackage{csvsimple}
\usepackage{subcaption}
\usepackage{multicol}
\usepackage[boxed]{algorithm2e}	
\usepackage{listings}
\usepackage{autobreak}

\usepackage{pgfplots}
\pgfplotsset{compat=1.14}
\usetikzlibrary{arrows.meta}

\makeatletter
\newcommand{\customlabel}[2]{%
   \protected@write \@auxout {}{\string \newlabel {#1}{{#2}{\thepage}{#2}{#1}{}} }%
}
\makeatother

\makeatletter
\newcommand{\customlabelS}[2]{%
   \protected@write \@auxout {}{\string \newlabel {#1}{{#2}{\thepage}{#2}{#1}{}} }%
   \hypertarget{#1}{#2}
}
\makeatother

\allowdisplaybreaks

\usepackage{array}
\newcolumntype{L}[1]{>{\raggedright\let\newline\\\arraybackslash\hspace{0pt}}m{#1}}
\newcolumntype{C}[1]{>{\centering\let\newline\\\arraybackslash\hspace{0pt}}m{#1}}
\newcolumntype{R}[1]{>{\raggedleft\let\newline\\\arraybackslash\hspace{0pt}}m{#1}}

\DeclareMathOperator*{\argmin}{argmin}

\def\boxl{\begingroup\setlength{\arraycolsep}{0pt}\begin{array}{|c|}\hline}

\def\boxr{\\ \hline\end{array}\endgroup}

\pgfplotsset{casefigmain/.style=
	{nodes near coords, mark options={scale=0.3},
	 scatter/classes={I={blue},O={red}},scatter,
	 only marks,point meta=explicit symbolic,
	 table/header = false, table/meta index = 2}}

\def\fig#1{Figure \ref{fig:#1}}

\vldbTitle{In-Machine-Learning Database}
\vldbAuthors{Len Du}
\vldbDOI{https://doi.org/10.14778/xxxxxxx.xxxxxxx}
\vldbVolume{xx}
\vldbNumber{x}
\vldbYear{2020}

\title{In-Machine-Learning Database:  \\
Reimagining
Deep Learning with Old-School SQL 
}

\numberofauthors{1}
\author{
\alignauthor
Len Du\\
\affaddr{Australian National University}\\
\email{len.du@anu.edu.au}
}

\def\headindoc*{ }

\def\beforebib*{
	\balance
	\bibliographystyle{abbrv}
}

\def\Section#1{\section{#1}}

\def\goodsinglewidth*{0.6\textwidth}
\def\smallsinglewidth*{0.4\textwidth}
\def\gooddoublewidth*{0.5\textwidth}
\def\twoorthree*{0.33\textwidth}

\def\citep#1{\cite{#1}}
\def\citet#1{\cite{#1}}

\def\ttt#1{\texttt{#1}}

\begin{document}

\headindoc*

\maketitle

\begin{abstract}
In-database machine learning has been very popular, almost being a cliche.
However, can we do it the other way around?
In this work, we say “yes” by applying plain old SQL to deep learning,
 in a sense implementing deep learning algorithms with SQL.

Most deep learning frameworks, as well as generic machine learning ones,
share a de facto standard of 
multidimensional array operations, underneath
fancier infrastructure such as automatic differentiation.
As SQL tables can be regarded as generalisations of (multi-dimensional) arrays,
we have found a way to express common deep learning
operations in SQL, 
encouraging a different way of thinking
and thus potentially novel models.
In particular, one of the latest trend in deep learning was the introduction of
sparsity in the name of graph convolutional networks, whereas we take sparsity
almost for granted in the database world.

As both databases and machine learning involve transformation of datasets,
we hope this work can inspire further works utilizing 
the large body of existing
wisdom, algorithms and technologies in the database field
to advance the state of the art in machine learning, rather
than merely integerating machine learning into databases.

\end{abstract}

\Section{Introduction}

Both machine learning and databases obviously
involve transformation of (or computation over) 
collections of numbers.
Combining the two fields is then an obvious conclusion.
But the way of such fusion seems to have been unilateral.
Much more effort has been spent towards providing machine
learning capabilities in a database context, 
or so-called “In-Database Machine Learning” \citep{MLog},
compared to integeration in the opposite direction,
which we call “In-Machine-Learning Database”.

We speculate that the connotation of databases has been more towards “systems”
than towards algorithms, compared to 
that of machine learning, making it seemingly more natural to 
apply the latter to the former.
Modern machine learning, in particular deep learning,
has been growing into expansive software systems as well,
which suggests us to seriously consider the reverse.

In this work we get back at the basic (or not so basic) notion
of “transforming collections of numbers” and try 
substituting the typical operations in machine learning
with the most prominent tool in databases, i.e. SQL,
to see whatever novel we can find under this different perspective.

\Section{Related Work}

In this section, we review some representative works connecting the two fields of machine learning (in particular, deep learning)
and databases, 
so that we can position this work properly in the whole data science landscape.
 In particular, reviewing these works helps us with a bird's-eye view of why the
relational model, having been ubiquitous in databases since the beginning of 
the field, should still interest those at the tip of deep learning research.

\subsection{Machine Learning in Databases}

MADlib \citep{MADlib} is probably the apex of the classical approach where
machine learning subroutines are provided as black-boxes in SQL.
MADlib also focuses on conventional machine learning rather than deep learning.

SciDB \citep{SciDB-1,SciDB-2} substitutes relational tables with multidimensional arrays.
In-database linear algebra and analytics can then be added,
resulting in a crossover between a numerical library and a database.

Tensor-Relational Model	\citep{TensorDB} is an elaborated treatise on
the role of multidimensional arrays in relational databases. 

MLog \citep{MLog} provides a domain-specific language designed
for deep learning. The MLog language is
integerated into RDBMS by mixing with SQL.
It operates on multidimensional arrays (tensors) rather than relational tables.
The implementation compiles MLog into TensorFlow\citep{tensorflow} programs.

In \citet{in-db-grad}, array operations, automatic
differentiation and gradient descent
are implemented via SQL extensions.

\citet{db-meets-dl} envisioned some possible ways to enhance database 
functionality with deep learning, 
beyond ease of access of deep learning in databases.

A strong argument favouring in-database machine learning is that 
databases are often mature distributed systems, so distributed machine learning
would supposedly require little extra effort on the user in a database setting.
\citep{in-db-ml-dist} explores such a setting.

\subsection{Database Functionality in General Machine Learning Settings}

Despite claimed as “an in-database framework”, 
AIDA \citep{RDBMS-friendly,AIDA} provides
a client interface to a SQL server
in the Python language, which is the de facto standard
in the machine learning world,
AIDA also shifts some of the computation 
to the server side, or more precisely,
a Python interpreter embedded in the database server.
So AIDA is best understood as implementing (low-level 
computation of) machine learning
in a database, and then providing the augmented database
to machine learning to the user.

ML2SQL \citep{ML2SQL,MLearn} compiles
a unified declarative domain-specific language to both database operations 
in SQL and ML-style array operations in python.

SystemML \citep{SystemML} and its successor SystemDS \citep{SystemDS}
also provide a unified language, but they use non-relational databases.

TensorLog \citep{TensorLog} implements probabilistic logic, essential
to probabilistic databases, over typical deep learning infrastructure.

In addition to machine learning in databases, \citet{db-meets-dl}
also	
envisions providing system-level facilities and distributed computation
developed in the database community to deep learning.

Finally, the Pandas \citep{Pandas} library familiar to data scientist 
already provides
some essential relational functionalities 
such as \texttt{JOIN} and \texttt{SELECT}.

\subsection{Neural Networks Designed for Relational Models}

There are also neural networks specifically designed for learning relations.
\citet{rl-exp-cmp} summerizes very well
the effort in this regard
before the "deep learning takeover", including
Graph Neural Networks (GNN) \citet{gnn},
and Relational Neural Networks \citet{rel-nn}.

In the more recent surge of deep learning,
\citet{rel-deepmind} explores a general deep learning architecture whose outputs
are relations, with applications to understanding scenes.
\citet{rel-deepmind-rnn} combines relational reasoning with recurrent neural networks.
\citet{rel-deepmind-rl} further applies the architecture in \citet{rel-deepmind} to complex reinforcement learning tasks.
\citet{RelNN} employs a modified logistic regression over hidden layers to learn relations.
Lifted Relational Neural Networks \citep{rel-nn-lifted}
combines first-order logic with neural networks to learn relational structures.

Note that all of these models \emph{designed to learn relations}, 
while worth mentioning, overlap little
with our claim that relational-model-based SQL can be
used as building blocks for \emph{general deep learning}.

\subsection{Relational Models versus Graph Convolutional Networks}

Even being the "fanciest of the fanciest" topic in Machine Learning, 
Graph convolutional network \citep{semi-gcn} (GCN)
can't escape the link to relational models \citep{rel-gcn}.
\citet{rel-deepmind}
advocates
relating relational models  and graph convolutional networks, as well as deep learning in general,
 with extensive review.

One interesting fact about GCNs is that GPUs no longer make the usual vast speedups. Even without consideration of relations, GPUs failed to accelerate beyond one order of magnitude 
\citep{semi-gcn}.
We speculate that it is something inherent given  the underlying  sparsity,
posing the same challenge to deep learning and databases alike.

Apparently, edges of graphs are relations.
But relations are not always edges -- they could be hyperedges!
From the point of view  of  the “ relational”  people, it is really a no-brainer that we could have hypergraph variants
of GCNs.
\citet{hgnn}  discusses them without addressing relational models while
\citet{hgcn} and \citet{dhgnn} address both relational models and hypergraph neural networks.

\def\SecRef#1{Section \ref{sec:#1}}

\def\relatedwork#1{\subsection{#1}}

\Section{Architecture}

In this part, we give a big picture of our proposed way of doing deep learning with SQL.
While the meaning of deep learning may not be exact enough to prevent intentionally 
creating a counterexample to
our arguments, actual instances of deep learning almost universally follow
the structures described here, at least in a practical, computational sense.

\subsection{ The (Usual) Way of Deep learning }

A deep learning model can usually be regarded as a scalar function $\mathcal f(\mathcal D,\mathcal P)$,
where $\mathcal D$ denotes a set of inputs (data) and $\mathcal P$ denotes the model parameters.
Our hypothetic goal is to find 
\begin{align*}
	\argmin\limits_{\mathcal P} \mathcal f (\mathcal D_{0},\mathcal P)
\end{align*}
where $\mathcal D_0$ can be interpreted as either the set of all possible inputs or a test set.

The global minimization is usually intractable.
So the learning process involves some iterative optimization involving the gradients
$\dfrac{\partial \mathcal f(\mathcal D,\mathcal P)}{\partial \mathcal P}$.
The iterative optimization takes the steps shown in Algorithm \ref{typical-dl}.
\begin{algorithm}
Load training set as $\mathcal D$ \;
Randomly initialize $\mathcal P$ \;
\While{not meeting stopping criteria}{
	(1) Evaluate	$\mathcal f(\mathcal D,\mathcal P)$ \; 
	(2) Evaluate	$\dfrac{\partial \mathcal f(\mathcal D,\mathcal P)}{\partial \mathcal P}$ (taken care of by automatic differentiation; not necessarily mathematically precise) \;
		(3) Update $\mathcal P$ with a new value computed with the old $\mathcal P$ and $\dfrac{\partial \mathcal f(\mathcal D,\mathcal P)}{\partial \mathcal P}$
	(may carry over state from previous iterations)
		\;
}
\caption{Typical deep learning control flow \label{typical-dl}}
\end{algorithm}
 This is the most basic pattern.  Some deep learning
 algorithms follow alternative versions.
  In real world scenarios, it is often only possible to train with stochastic
 gradient descent which follows the general pattern outlined in Algorithm \ref{typical-sgd},
 where only a subset of $\mathcal D$ is picked for training each iteration.
 \begin{algorithm}
Load $\mathcal D$ ; Initialize $\mathcal P$ \;
\While{not meeting stopping criteria}{
	(1) Evaluate $\mathcal f(\mathcal D',\mathcal P)$ where $\mathcal D'\in\mathcal D$ (chosen per iteration) \;
	(2) Evaluate $\dfrac{\partial \mathcal f(\mathcal D,\mathcal P)}{\partial \mathcal P}$  \;
	(3) Update $\mathcal P$ \;
}
\caption{Deep learning control flow that stochastic gradient descent uses\label{typical-sgd}}
\end{algorithm}
 A recurrent neural network 
predicting the next symbol given a prefix
would be trained with Algorithm \ref{typical-rnn},
 in a self-supervised fashion,
 where the samples are encoded one megasequence of vectors $\mathcal S$.
\begin{algorithm}
Load $\mathcal S$ ; Initialize $\mathcal P$ \;
\While{not meeting stopping criteria}{
	(1) Evaluate $\mathcal f(\mathcal S,\mathcal P)$: \\
	(1a) Initialize hidden states $\mathcal H$ ( typically with zeroes )\; 
	(1b) \For{ $\mathcal S'$ (embeddings of) sub-sequence of $\mathcal S$ }
	{
		(1b1) Evaluate network output (embeddings of predicted sub-sequence) $\mathcal S''$ 
	         and update hidden states $\mathcal H$ with
		      $(\mathcal S'',\mathcal H)\leftarrow f_1(S',\mathcal H,\mathcal P)$ \;
		(1b2) Evaluate per-sub-sequence loss $f_2(\mathcal S'',\mathcal S)$ 
		      by comparing prediction $\mathcal S''$ and the corresponding (embeddings of) 
				sub-sequence of $\mathcal S$ \;
	} {}
	(1c) compute total loss $\mathcal f(\mathcal S,\mathcal P)$  by summing or averaging all per-sub-sequence losses \;
	(2) Evaluate $\dfrac{\partial \mathcal f(\mathcal S,\mathcal P)}{\partial \mathcal P}$  \;
	(3) Update $\mathcal P$ \;
}
\caption{Control flow for training recurrent neural networks\label{typical-rnn}}
\end{algorithm}

 Even unconventional uses of deep learning are not so unconventional in terms of
 control flows. Neural style transfer \citep{Gatys2016} follows the pattern in Algorithm \ref{typical-nst},
 essentially just switching the argument in Algorithm \ref{typical-dl} from 
 $\mathcal P$ to $\mathcal D$.
\begin{algorithm}
Load content image $\mathcal I'$ and style image $\mathcal I''$ \;
Initialize image $\mathcal I$ ( maybe randomly but usually $\mathcal I \leftarrow \mathcal I'$)\; 
Load (pre-trained) $\mathcal P$ \;
\While{not meeting stopping criteria}{
	(1) Evaluate $\mathcal f(\mathcal I,\mathcal I',\mathcal I'',\mathcal P)$ \;
	(2) Evaluate $\dfrac{\partial \mathcal f(\mathcal I,\mathcal P)}{\partial \mathcal I}$  \;
	(3) Update $\mathcal I$ \;
}
\caption{Control flow for neural style transfer\label{typical-nst}}
\end{algorithm}
Note that the content image $\mathcal I'$ and style image $\mathcal I''$ 
 are never modified once loaded. In practice, it is often possible
 to leave out $\mathcal I'$ in the iterative optimization altogether 
 so long as $\mathcal I'$ is used as the initial value of $\mathcal I$ 
 \citep{nst-ablation}.
Adversial example generation, like fast gradient sign attack \citep{fgsa},
works in a very similar way by perturbing $\mathcal D$ to maximize
$\mathcal f(\mathcal D,\mathcal P)$ in Algorithm \ref{typical-dl} rather than perturbing $\mathcal P$ to minimize it.
 The phenomenal generative adversarial network (GAN) pipes two
 ordinary networks with parameter sets $\mathcal P_1$ and $\mathcal P_2$
 together and run two optimizations in lockstep as shown in Algorithm \ref{typical-gan}.
\begin{algorithm}
Initialize $\mathcal P_1$, $\mathcal P_2$\;
Load true samples $\mathcal D$\;
\While{not meeting stopping criteria}{
	(1) Evaluate $ \mathcal f(\mathcal D,\mathcal N,\mathcal P_1,\mathcal P_2) 
	= f_2(\mathcal D,f_1(\mathcal N,\mathcal P_1),\mathcal P_2) $ 
	where $\mathcal N$ is some kind of noise \;
				(2) Evaluate $\dfrac{\partial \mathcal f(\mathcal D,\mathcal N,\mathcal P_1,\mathcal P_2)}{\partial \mathcal P_1}$ and $\dfrac{\partial \mathcal f(\mathcal D,\mathcal N,\mathcal P_1,\mathcal P_2)}{\partial \mathcal P_2}$ \;
	(3) Update $\mathcal P_1$ (maximizing) and $\mathcal P_2$ (minimizing) \\ according to respective gradients\;
}
\caption{Control flow for generative adversarial networks \label{typical-gan}}
\end{algorithm}
 Function $f_1$ along with parameters $\mathcal P_1$ is the so-called generator network producing
 “fake” samples given noise as input, 
 while  the discriminator network with parameters $\mathcal P_2$ trying work out a 
 score for each of
 both these “fake” samples and the
 “real” ones given as the training set.
 Then, one number representing how well the scores separate the
 two types of samples is summarized from the scores.
 Finally the two sets of parameters are optimized with respect to this number, 
 albeit with opposite signs.
 Surely the order of steps (2) and (3) does not matter.

While there could be other ways to code a deep learning  program,
the pattern is quite clear.
The control flows of deep learning programs are relatively simple,
whereas the bulk of the effort are distributed to the design of the models,
 manifesting primarily in Step (1) of each example,
 among the 3 major steps conveniently partitioned out of the main loops.

 As for Step (2) and (3), there is a separation of concern here.
Pragmatically, a major breakthrough that enabled the explosive 
progress of deep learning is the automatic differentiation.
While still an active field of research, development
of new deep learning models can be separated from 
studying automatic differentiation (Step (2)) itself.
We can mix and match different flavors of control flows
with different optimization algorithms, or more precisely, different
update strategies (Step (3)).
While some combinations work better than  others,
in general inventors of new deep learning models
do not concern themselves with which update strategy
to pick until tuning the performance of the model.

\subsection{Tensor to Relations and back again}

Modern deep learning infrastructure has been almost universally built upon array-oriented programming paradigms.
In this work, we concern ourselves with expressing the deep learning model $\mathcal f$
in (a very limited set of) SQL.

\Section{DL-in-SQL by Examples}

While we are far from a formal proof that SQL can express every possible deep learning model
because of the obvious lack of precise definition of the latter, 
we nevertheless demonstrate how the bread-and-butter constructs of deep learning
can be expressed in SQL.

Here, we use an example deep learning task in stark contrast to 
typical database-related ones,
to demonstrate that our architechture is really
geared towards deep learning in general. We will introduce how frequent layers can be cast into SQL along the way.
Without further ado, we begin the demonstration.

\subsection{Image Classifier Convolutional Networks}

\begin{figure}
\includegraphics[width=\linewidth]{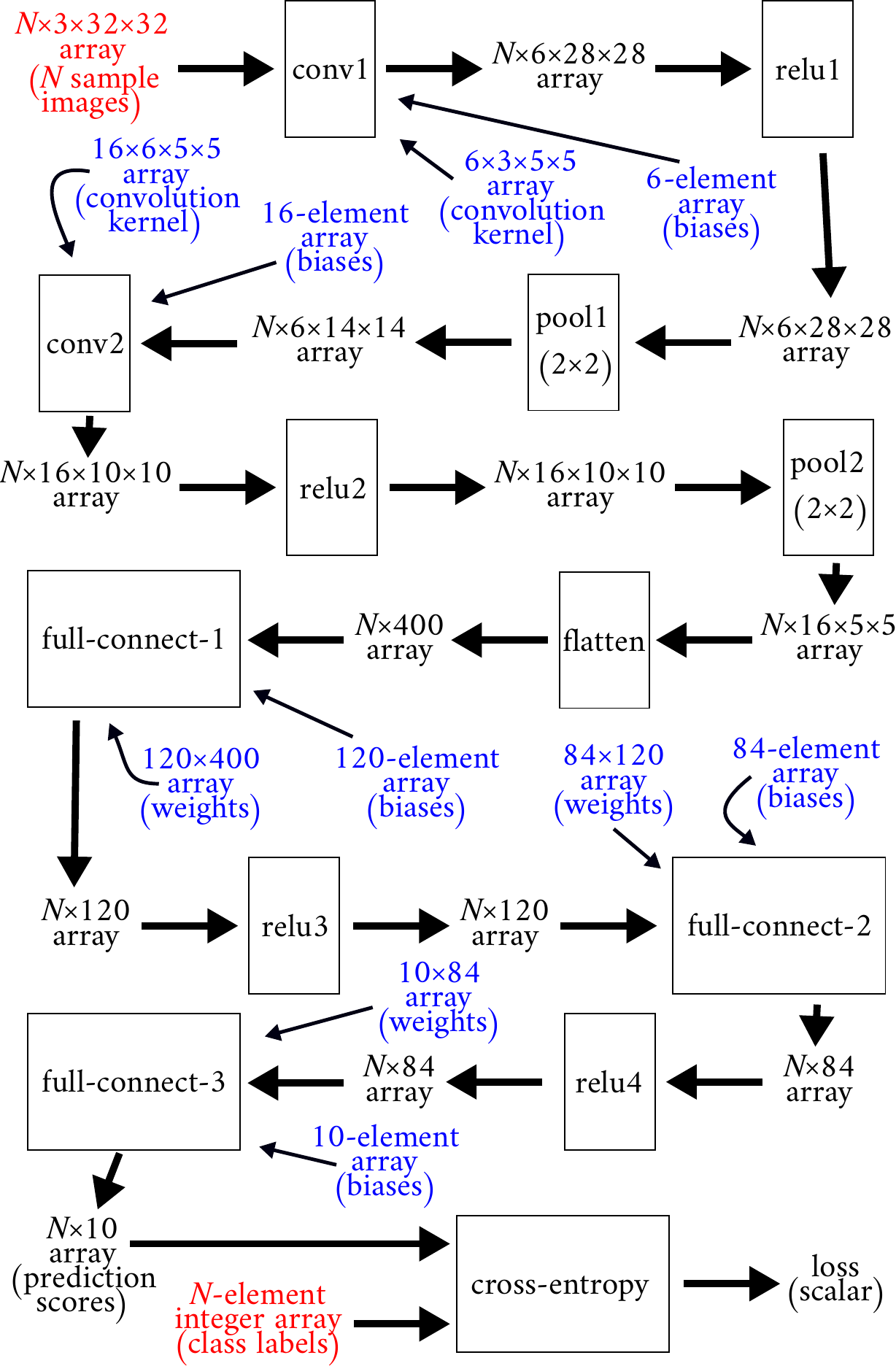}
\caption{Structure of the CNN for classifying images.
Steps of 
computation are \framebox{framed}, while their inputs and outputs
 are not.
Data (parts of $\mathcal D$) are marked in \textcolor{red}{red},
while parameters (parts of $\mathcal P$) are marked in \textcolor{blue}{blue}.
\label{fig:CNN-Demo}}
\end{figure}

In this example, we demonstrate how to specify
a convolutional neural network for computer vision in SQL.
The model takes $N$ sample images together as input,
where $N$ varies depending on how the model is used.
The model is fixed for 10 classes, and $32\times 32$ RGB images.
Computationally, the neural network displays the structure
illustrated in \fig{CNN-Demo}.
To make things crystal clear,
we have drawn all parameters of the network explicitly,
so each “step of computation” in a box does not contain any states.
This is quite different from many illustrations found elsewhere.
For instance, in deep learning jargon,
the first “convolutional layer” would conceptually include both “conv1” in the box
and the two parameter arrays (kernel and biases) marked in blue,
usually not shown explicitly in diagrams.

Now, we essentially need to express the boxed
steps of computation in SQL, with {data $\mathcal D$ in red} and 
{paramters $\mathcal P$ in blue} given as SQL tables.

First and foremost, let us see what we can do with the convolution step
\framebox{conv1}. In the SQL context, we provide the 4-dimensional ($N\times 3\times 32\times 32$) array of $N$ sample images 
as
a relation \texttt{samples} with the following 5 columns.
\begin{center}
\texttt{\begin{tabular}{r|l}
image, channel, r, c&INTEGER \\ val&REAL
\end{tabular}}
\end{center}
The names and types of the columns should be quite 
self-explanatory.
The column \ttt{image}
refers to indices in the first dimension of the original 4D array,
taking the values from $0$ to $(N-1)$ (inclusive).
Similarly, the column \ttt{channel} refers to which one of the
three (RGB) channels (2nd dimension), while \ttt{r} and \ttt{c} refer to 
which row (3rd dimension) and which column (4th dimension) respectively.
The column \ttt{val} stores the actual values in the array, obviously.

Similarly, the the $6\times 3\times 5\times 5$ array of the convolution kernel is 
presented as a relation \ttt{conv1\_weight} with 5 columns.
\begin{center}
\texttt{\begin{tabular}{r|l}
out\_channel, in\_channel, r, c&INTEGER\\ weight&REAL
\end{tabular}}
\end{center}
And the biases to the convolutional layer corresponds to a 2-column relation \ttt{conv1\_bias}.
\begin{center}
\texttt{\begin{tabular}{r|l}
out\_channel&INTEGER\\bias&REAL
\end{tabular}}
\end{center}

With the input relation ready,
we can execute the computation of \framebox{conv1} with \ttt{CREATE TABLE} commands.
We do this in two steps. 
Firstly, we put the results of the \emph{convolution} itself 
into \ttt{conv1\_unbiased}. 
\begin{lstlisting}[language=SQL]
CREATE TABLE conv1_unbiased AS 
	SELECT 
	image, 
	out_channel AS channel, 
	samples.r-conv1_weight.r AS r1, 
	samples.c-conv1_weight.c AS c1, 
	SUM(val * weight) AS val 
	FROM samples, conv1_weight 
	WHERE 
	channel = in_channel AND 
	r1 BETWEEN 0 AND 32-5 AND 
	c1 BETWEEN 0 AND 32-5 
	GROUP BY image,channel,r1,c1;
\end{lstlisting}
Then we apply the biases to get \ttt{conv1\_out}.
\begin{lstlisting}[language=SQL]
CREATE TABLE conv1_out AS 
	SELECT 
	image, channel, r1 AS r, c1 AS c, 
	val + bias AS val 
	FROM conv1_unbiased,conv1_bias 
	WHERE channel = out_channel;
\end{lstlisting}

The ReLU layer \framebox{relu1} is embarrassingly simple to compute.
\begin{lstlisting}[language=SQL]
CREATE TABLE relu1_out AS
	SELECT 
	image, channel, r, c, MAX(0,val) AS val
	FROM conv1_out
\end{lstlisting}

Executing max-pooling (\framebox{pool1}) is also straight-forward.
\begin{lstlisting}[language=SQL]
CREATE TABLE pool1_out AS
	SELECT
	image, channel, r/2 AS r, c/2 AS c, 
	MAX(val) AS val
	FROM relu1_out GROUP BY image,channel, r, c;
\end{lstlisting}

Now the remaining computation steps up till \framebox{flatten} 
are almost identical to what we have listed except for
table names and array dimensions.
We skip them to assume having evaluated the
output of \framebox{pool2}.
\framebox{flatten} is almost as simple as ReLU.
\begin{lstlisting}[language=SQL]
CREATE TABLE flatten_out AS
	SELECT 
	image, (channel * 5 + r)*5+c AS i, val 
	FROM pool2_out;
\end{lstlisting}

A fully-connected layer like \framebox{fc1} is treated 
just like a convolution layer, only simpler.
The weights and biases are put in the SQL context 
as \verb|fc1_weight| with
\begin{center}
\texttt{\begin{tabular}{r|l}
out\_dim, in\_dim&INTEGER\\weight&REAL
\end{tabular}}
\end{center}
and \verb|fc1_bias| with
\begin{center}
\texttt{\begin{tabular}{r|l}
out\_dim&INTEGER\\bias&REAL
\end{tabular}}~.	
\end{center}
Then we can compute \verb|fc1_out|
by applying weights and biases in two 
consecutive steps.
\begin{lstlisting}[language=SQL]
CREATE TABLE fc1_unbiased AS 
	SELECT 
	image, out_dim AS i, 
	SUM(val * weight) AS val 
	FROM flatten_out, fc1_weight 
	WHERE i = in_dim GROUP BY image, out_dim; 
CREATE TABLE fc1_out AS 
	SELECT image, i, val + bias AS val 
	FROM fc1_unbiased,fc1_bias 
	WHERE i=out_dim;
\end{lstlisting}

At this point the computation in SQL
up to the output of \framebox{full-connect-3}
should be clear.
At this stage we could claim that 
we have specified the \emph{neural network per se}.
It is enough for executing inference.
However for training, we still have to show
how to compute \framebox{cross-entropy}.

Cross-entropy loss for one sample of computed 
label weights $\mathbf x$ 
whose
correct label is $l$ is given by
\begin{align*}
\text{loss} (\mathbf x,l)=-x_{l}+\log\limits (\sum _{j} \text{exp} (x_{j}))~.
\end{align*}
And we choose to compute the loss
over the $N$ samples as the mean 
over each sample.
Assume the output of \framebox{full-connect-3}
to be \verb|fc3_out|.
First we compute the right side of 
“+” with
\begin{lstlisting}[language=SQL]
CREATE TABLE x_ent_losses_r AS
	SELECT image, LOG(SUM(EXP(val))) AS r 
	FROM fc3_out GROUP BY image;
\end{lstlisting}
where \texttt{LOG()} and 
\texttt{EXP()} are obviously
(element-wise)
natural logarithm and exponentiation.
The left hand side is just selecting
one of 10 element, listed as follows
list it for clarity.
\begin{lstlisting}[language=SQL]
CREATE TABLE x_ent_losses_l AS
	SELECT fc3_out.image,-val AS l 
	FROM fc3_out, labels 
	WHERE 
	fc3_out.image = labels.image AND i = label;
\end{lstlisting}
Then we combine both sides to
get loss for each image
\begin{lstlisting}[language=SQL]
CREATE TABLE x_ent_losses AS
	SELECT image, l+r AS val
	FROM 
	x_ent_losses_l NATURAL JOIN x_ent_losses_r;
\end{lstlisting}
and then finally the average loss
\begin{lstlisting}[language=SQL]
CREATE TABLE x_ent_loss AS
	SELECT SUM(val)/COUNT(val)
	FROM x_ent_losses;
\end{lstlisting}
as an 1-row table.
Now we have finished 
specifying a deep learning model
entirely in SQL.

\subsection{Graph Convolutional Network}

Now let us see a baisc Graphan  
Convolutional Network (GCN) setup
in SQL.
This GCN is adapted from 
a simplified version of that
in \citet{gcn_tut}, as 
a very neat tutorial
provided by the same 
author \citet{pygcn}.
We further remove the dropout 
to simplify things,
which moderately increases the computation
load and slightly increases overfitting
while not changing major results.

\begin{figure}
\includegraphics[width=\linewidth]{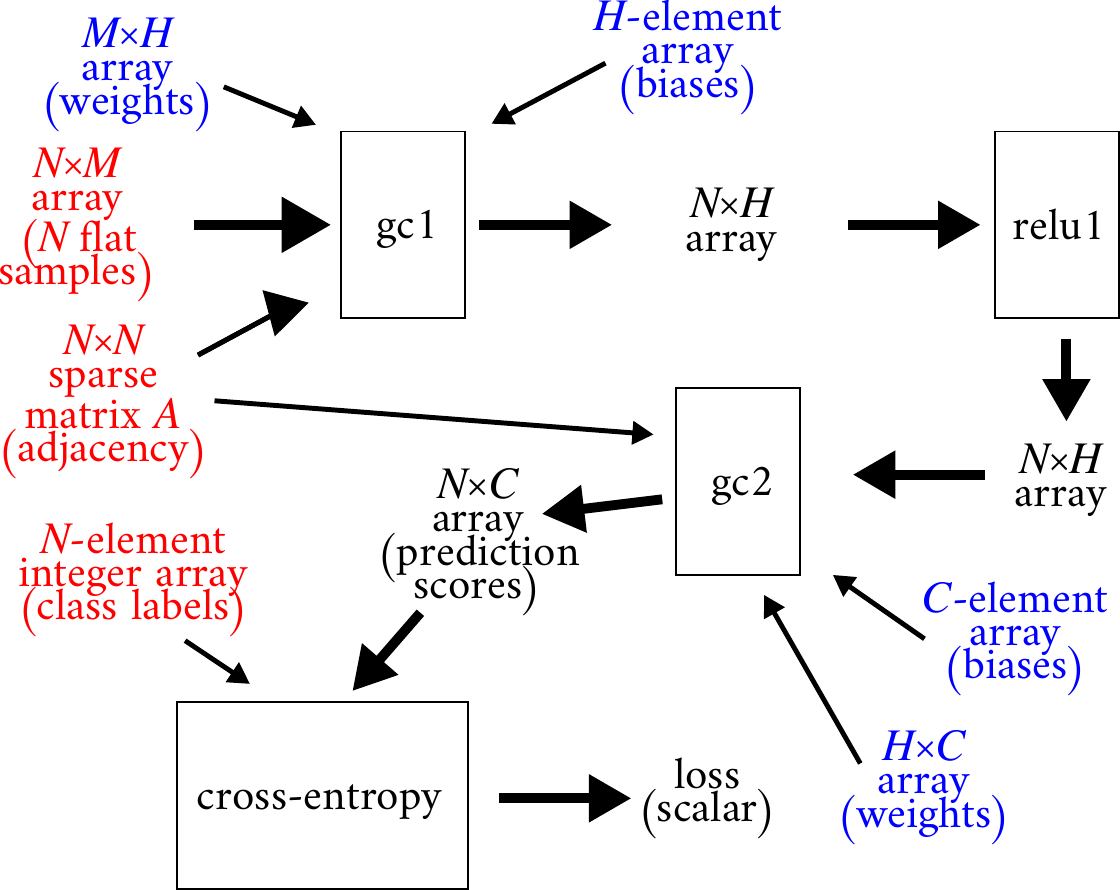}
\caption{Structure of the simple examplar 
Graph Convolutional Network.
Steps of 
computation are \framebox{framed}, while their inputs and outputs
 are not.
Data (parts of $\mathcal D$) are marked in \textcolor{red}{red},
while parameters (parts of $\mathcal P$) are marked in \textcolor{blue}{blue}.
\label{fig:GCN-Demo}}
\end{figure}

The structure of the whole forward computation
is as shown in \fig{GCN-Demo}.
This time we assume $N$ samples
of $M$ features to be classified
into $C$ classes, with 
one hidden layer of size $H$ 
in-between.
This structure is actually 
much simpler than the previous 
example, the
only really new things being the
Graph Convolutional layers
(\framebox{gc1} and
\framebox{gc2})
and the accompanying 
($N\times N$)
adjacency matrix 
$A$. 
So we will focus on them.

From a computational point-of-view,
what the Graph Convolutional layer can be 
considered as
two consecutive matrix multiplications
plus biasing,
despite named “convolutional” layers.
That is, the computation before biasing 
can be simply expressed as
$
A X W~,
$
where $X$ denotes the input of the layer
and $W$ denotes the weights.
Take \framebox{gc1} for example,
$X$ is an $N\times M$ matrix while
while $W$ is $M\times H$.
Then we can add the biases with
\begin{align*}
Y_{i,j}=(A X W)_{i,j}+B_{j}~,
\end{align*}
for all $i\in  \{ 1..N \} ,j\in  \{ 1..H \} $,
$B$ being the biases of the layer
\framebox{gc1}.

Now continuing with \framebox{gc1},
we assume the following tables 
in the SQL world.
\begin{lstlisting}[language=SQL]
samples(i INTEGER, j INTEGER, val REAL);
gc1_w  (i INTEGER, j INTEGER, weight REAL);
gc1_b  (i INTEGER, bias REAL);
adj    (i INTEGER, j INTEGER, val REAL);
\end{lstlisting}
Drawing from previous experience
from fully-connected layers,
We can easily reproduce the 
above forward computation in SQL
with 
\begin{lstlisting}[language=SQL]
CREATE TABLE gc1_mid AS
	SELECT
	samples.i AS i, gc1_w.j AS j,
	SUM(val * weight) AS val
	FROM samples, gc1_w
	WHERE samples.j == gc1_w.i
	GROUP BY samples.i, gc1_w.j;
\end{lstlisting}
for the intermediate matrix product 
$XW$,
and subsequently
\begin{lstlisting}[language=SQL]
CREATE TABLE gc1_out AS
	SELECT
	adj.i AS i, gc1_mid.j AS j,
	SUM(adj.val * gc1_mid.val) + bias AS val
	FROM adj, gc1_mid, gc1_b
	WHERE
	adj.j == gc1_mid.i AND gc1_mid.j == gc1_b.i
	GROUP BY adj.i, gc1_mid.j;
\end{lstlisting}
for the whole biased output.

The symmetric adjacency matrix $A$ has zeroes for pairs of
vertices without an edge in-between and
non-zeroes for those connected by an edge.
Furthermore, the adjacency matrix is row-normalized
from a typical adjacency matrix of 0's and 1's.
That is, each row sums to either 1 if there are any
edges on the vertex or 0 if the vertex is complete
isolated.
And so does each column because of symmetry.

The adjacency matrix here is usually a sparse matrix.
Or to put it another way, 
the sparsity is essential
to its practical effectiveness,
which in turn was probably an important
precondition to its current
popularity.
Yet in the SQL world 
we do not treat sparsity as something special.
Actually we take 
sparsity for granted.
While how to create
an implementation 
running as fast as array-based
deep-learning frameworks
is no easy task,
we already have 
battle-hardened
semantics and standards
in the database world.

\begin{lstlisting}[language=SQL]
\end{lstlisting}

\Section{Discussion}

The notion that databases provides the data 
for dedicated machine learning components to work on
has been seldom questioned when unifiying 
the two fields.
But we may as well go to the extent that  we
build machine learning programs in terms of
 databases.

 In general, we could consider databases  as the
 most feature-rich deep learning framework in the future.
 We look at databases and we know what can be added to
 deep learning.

 One often overlooked fact is that datasets are processed as
 arrays, implying an order while they are not supposed to.
 Relational algebra and subsequently SQL take care of this
 automatically.

 The biggest challenge for  implementation would be to port automatic differentation
 to relational algebra. However it could also be implemented
 over yet another layer of flat array framework with automatic
 differentiation, treated as some kind of linear memory to sidestep
 automatic differentiation from the ground up.

Databases have been dealing with sparse data to begin with.
While directly run deep learning in database engine
may not be competitive as 
random access too much to be fast at batch processing,
we can certainly continue to bring experience
in database to machine learning
for a very long time to come.
Perhaps when new deep learning models are proposed,
the machine learning researchers will find the 
database community waiting for them.

\beforebib*
\bibliography{imldb0,dl-cv,gcn}

\begin{thebibliography}{10}

\bibitem{tensorflow}
M.~Abadi, A.~Agarwal, P.~Barham, E.~Brevdo, Z.~Chen, C.~Citro, G.~S. Corrado,
  A.~Davis, J.~Dean, M.~Devin, S.~Ghemawat, I.~Goodfellow, A.~Harp, G.~Irving,
  M.~Isard, Y.~Jia, R.~Jozefowicz, L.~Kaiser, M.~Kudlur, J.~Levenberg,
  D.~Man\'{e}, R.~Monga, S.~Moore, D.~Murray, C.~Olah, M.~Schuster, J.~Shlens,
  B.~Steiner, I.~Sutskever, K.~Talwar, P.~Tucker, V.~Vanhoucke, V.~Vasudevan,
  F.~Vi\'{e}gas, O.~Vinyals, P.~Warden, M.~Wattenberg, M.~Wicke, Y.~Yu, and
  X.~Zheng.
\newblock {TensorFlow}: Large-scale machine learning on heterogeneous systems,
  2015.
\newblock Software available from tensorflow.org.

\bibitem{SystemDS}
M.~Boehm, I.~Antonov, M.~Dokter, R.~Ginthoer, K.~Innerebner, F.~Klezin,
  S.~Lindstaedt, A.~Phani, and B.~Rath.
\newblock Systemds: A declarative machine learning system for the end-to-end
  data science lifecycle.
\newblock 09 2019.

\bibitem{SystemML}
M.~Boehm, M.~W. Dusenberry, D.~Eriksson, A.~V. Evfimievski, F.~M. Manshadi,
  N.~Pansare, B.~Reinwald, F.~R. Reiss, P.~Sen, A.~C. Surve, et~al.
\newblock Systemml: Declarative machine learning on spark.
\newblock {\em Proceedings of the VLDB Endowment}, 9(13):1425--1436, 2016.

\bibitem{AIDA}
J.~V. D'silva, F.~De~Moor, and B.~Kemme.
\newblock Aida: abstraction for advanced in-database analytics.
\newblock {\em Proceedings of the VLDB Endowment}, 11(11):1400--1413, 2018.

\bibitem{RDBMS-friendly}
J.~V. D'silva, F.~{De Moor}, and B.~Kemme.
\newblock Making an {RDBMS} data scientist friendly: Advanced in-database
  interactive analytics with visualization support.
\newblock {\em {PVLDB}}, 12(12):1930--1933, 2019.

\bibitem{nst-ablation}
L.~Du.
\newblock How much deep learning does neural style transfer really need? an
  ablation study.
\newblock In {\em The IEEE Winter Conference on Applications of Computer Vision
  (WACV)}, March 2020.

\bibitem{hgnn}
Y.~Feng, H.~You, Z.~Zhang, R.~Ji, and Y.~Gao.
\newblock Hypergraph neural networks.
\newblock In {\em Proceedings of the AAAI Conference on Artificial
  Intelligence}, volume~33, pages 3558--3565, 2019.

\bibitem{Gatys2016}
L.~A. Gatys, A.~S. Ecker, and M.~Bethge.
\newblock Image style transfer using convolutional neural networks.
\newblock In {\em Proceedings of the IEEE Conference on Computer Vision and
  Pattern Recognition}, Jun 2016.

\bibitem{fgsa}
I.~J. Goodfellow, J.~Shlens, and C.~Szegedy.
\newblock Explaining and harnessing adversarial examples.
\newblock {\em arXiv preprint arXiv:1412.6572}, 2014.

\bibitem{MADlib}
J.~M. Hellerstein, C.~R{\'e}, F.~Schoppmann, D.~Z. Wang, E.~Fratkin,
  A.~Gorajek, K.~S. Ng, C.~Welton, X.~Feng, K.~Li, et~al.
\newblock The madlib analytics library.
\newblock {\em Proceedings of the VLDB Endowment}, 5(12), 2012.

\bibitem{dhgnn}
J.~Jiang, Y.~Wei, Y.~Feng, J.~Cao, and Y.~Gao.
\newblock Dynamic hypergraph neural networks.
\newblock In {\em Proceedings of the 28th International Joint Conference on
  Artificial Intelligence}, pages 2635--2641. AAAI Press, 2019.

\bibitem{TensorLog}
W.~W. C. F.~Y. Kathryn and R.~Mazaitis.
\newblock Tensorlog: Deep learning meets probabilistic databases.
\newblock {\em Journal of Artificial Intelligence Research}, 1:1--15, 2018.

\bibitem{RelNN}
S.~M. Kazemi and D.~Poole.
\newblock Relnn: A deep neural model for relational learning.
\newblock In {\em Thirty-Second AAAI Conference on Artificial Intelligence},
  2018.

\bibitem{TensorDB}
M.~Kim.
\newblock {\em TensorDB and tensor-relational model (TRM) for efficient
  tensor-relational operations}.
\newblock Arizona State University, 2014.

\bibitem{pygcn}
T.~N. Kipf.
\newblock https://github.com/tkipf/pygcn.

\bibitem{semi-gcn}
T.~N. Kipf and M.~Welling.
\newblock Semi-supervised classification with graph convolutional networks.
\newblock 2016.

\bibitem{gcn_tut}
T.~N. Kipf and M.~Welling.
\newblock Semi-supervised classification with graph convolutional networks.
\newblock {\em arXiv preprint arXiv:1609.02907}, 2016.

\bibitem{MLog}
X.~Li, B.~Cui, Y.~Chen, W.~Wu, and C.~Zhang.
\newblock Mlog: Towards declarative in-database machine learning.
\newblock {\em Proceedings of the VLDB Endowment}, 10(12):1933--1936, 2017.

\bibitem{Pandas}
W.~McKinney.
\newblock Data structures for statistical computing in python.
\newblock In S.~van~der Walt and J.~Millman, editors, {\em Proceedings of the
  9th Python in Science Conference}, pages 51 -- 56, 2010.

\bibitem{in-db-ml-dist}
S.~S. Sandha, W.~Cabrera, M.~Al-Kateb, S.~Nair, and M.~Srivastava.
\newblock In-database distributed machine learning: Demonstration using
  teradata sql engine.
\newblock {\em Proc. VLDB Endow.}, 12(12):1854–1857, Aug. 2019.

\bibitem{rel-deepmind-rnn}
A.~Santoro, R.~Faulkner, D.~Raposo, J.~Rae, M.~Chrzanowski, T.~Weber,
  D.~Wierstra, O.~Vinyals, R.~Pascanu, and T.~Lillicrap.
\newblock Relational recurrent neural networks.
\newblock In {\em Advances in neural information processing systems}, pages
  7299--7310, 2018.

\bibitem{rel-deepmind}
A.~Santoro, D.~Raposo, D.~G. Barrett, M.~Malinowski, R.~Pascanu, P.~Battaglia,
  and T.~Lillicrap.
\newblock A simple neural network module for relational reasoning.
\newblock In I.~Guyon, U.~V. Luxburg, S.~Bengio, H.~Wallach, R.~Fergus,
  S.~Vishwanathan, and R.~Garnett, editors, {\em Advances in Neural Information
  Processing Systems 30}, pages 4967--4976. Curran Associates, Inc., 2017.

\bibitem{gnn}
F.~Scarselli, M.~Gori, A.~C. Tsoi, M.~Hagenbuchner, and G.~Monfardini.
\newblock The graph neural network model.
\newblock {\em IEEE Transactions on Neural Networks}, 20(1):61--80, 2008.

\bibitem{rel-gcn}
M.~Schlichtkrull, T.~N. Kipf, P.~Bloem, R.~Van Den~Berg, I.~Titov, and
  M.~Welling.
\newblock Modeling relational data with graph convolutional networks.
\newblock In {\em European Semantic Web Conference}, pages 593--607. Springer,
  2018.

\bibitem{in-db-grad}
M.~Sch{\"u}le, F.~Simonis, T.~Heyenbrock, A.~Kemper, S.~G{\"u}nnemann, and
  T.~Neumann.
\newblock In-database machine learning: Gradient descent and tensor algebra for
  main memory database systems.
\newblock {\em BTW 2019}, 2019.

\bibitem{ML2SQL}
M.~E. Sch{\"u}le, M.~Bungeroth, D.~Vorona, A.~Kemper, S.~G{\"u}nnemann, and
  T.~Neumann.
\newblock Ml2sql - compiling a declarative machine learning language to sql and
  python.
\newblock In {\em EDBT}, 2019.

\bibitem{MLearn}
M.~Schüle, M.~Bungeroth, A.~Kemper, S.~Günnemann, and T.~Neumann.
\newblock Mlearn: A declarative machine learning language for database systems.
\newblock pages 1--4, 06 2019.

\bibitem{rel-nn-lifted}
G.~Sourek, V.~Aschenbrenner, F.~Zelezny, S.~Schockaert, and O.~Kuzelka.
\newblock Lifted relational neural networks: Efficient learning of latent
  relational structures.
\newblock {\em Journal of Artificial Intelligence Research}, 62:69--100, 2018.

\bibitem{SciDB-2}
M.~Stonebraker, P.~Brown, J.~Becla, and D.~Zhang.
\newblock Scidb: A database management system for applications with complex
  analytics.
\newblock {\em Computing in Science and Engg.}, 15(3):54–62, May 2013.

\bibitem{SciDB-1}
M.~Stonebraker, P.~Brown, A.~Poliakov, and S.~Raman.
\newblock The architecture of scidb.
\newblock In {\em Proceedings of the 23rd International Conference on
  Scientific and Statistical Database Management}, SSDBM’11, page 1–16,
  Berlin, Heidelberg, 2011. Springer-Verlag.

\bibitem{rel-nn}
W.~Uwents and H.~Blockeel.
\newblock Classifying relational data with neural networks.
\newblock In S.~Kramer and B.~Pfahringer, editors, {\em Inductive Logic
  Programming}, pages 384--396, Berlin, Heidelberg, 2005. Springer Berlin
  Heidelberg.

\bibitem{rl-exp-cmp}
W.~Uwents, G.~Monfardini, H.~Blockeel, M.~Gori, and F.~Scarselli.
\newblock Neural networks for relational learning: an experimental comparison.
\newblock {\em Machine Learning}, 82:315--349, 03 2011.

\bibitem{db-meets-dl}
W.~Wang, M.~Zhang, G.~Chen, H.~Jagadish, B.~C. Ooi, and K.-L. Tan.
\newblock Database meets deep learning: Challenges and opportunities.
\newblock {\em ACM SIGMOD Record}, 45(2):17--22, 2016.

\bibitem{hgcn}
N.~Yadati, M.~Nimishakavi, P.~Yadav, V.~Nitin, A.~Louis, and P.~Talukdar.
\newblock Hypergcn: A new method for training graph convolutional networks on
  hypergraphs.
\newblock In {\em Advances in Neural Information Processing Systems}, pages
  1509--1520, 2019.

\bibitem{rel-deepmind-rl}
V.~Zambaldi, D.~Raposo, A.~Santoro, V.~Bapst, Y.~Li, I.~Babuschkin, K.~Tuyls,
  D.~Reichert, T.~Lillicrap, E.~Lockhart, M.~Shanahan, V.~Langston, R.~Pascanu,
  M.~Botvinick, O.~Vinyals, and P.~Battaglia.
\newblock Deep reinforcement learning with relational inductive biases.
\newblock In {\em International Conference on Learning Representations}, 2019.

\end{thebibliography}

\end{document}